# A Reliable Fault Diagnosis Method Based on Belief Rule Base Consider Robustness Analysis

Mingyuan Liu[1], Dan Yin[1], Zongzong Wu[1]

[1] Central South University, Changsha, Hunan, P.R. China. 410083

**Abstract:** In equipment operation, the implementation of fault diagnosis is essential to ensure the continuity and safety of production equipment, improve operational efficiency and reduce maintenance costs. Since sensor readings are widely used for fault diagnosis, their reliability directly affects the results of fault diagnosis. A new fault diagnosis method is proposed to address the two problems of robustness assessment and robustness optimization of fault diagnosis models. For this purpose, a reliable fault diagnosis method based on a belief rule base (BRB) considering robustness analysis is proposed. Firstly, the robustness analysis of the BRB model is carried out systematically. Secondly, three robustness constraint strategies are proposed to optimize the robustness of the BRB fault diagnosis model. Finally, the effectiveness of the proposed model is verified by taking the fault diagnosis of WD615 diesel engine and Case Western Reserve University bearings as an example, and the experiments show that the proposed model improves both accuracy and robustness.

## 1 Introduction

In the field of engineering, equipment fault diagnosis is of great importance. With the advancement of industrial technology, equipment is gradually becoming highly integrated and complex, which increases the difficulty and importance of diagnosis. As a result, improving the reliability of troubleshooting has become a topic of widespread interest, aiming to ensure continuous operation of equipment, maximize productivity, and reduce operating costs [1][2].

Fault diagnosis serves as a crucial technique for identifying and addressing equipment malfunctions, relying on gathered observational data to ensure its diagnostic precision[3]. Traditionally, fault diagnosis methodologies are categorized into three types: knowledge-based, data-driven, and hybrid approaches that utilize both data and expert knowledge. Contemporary research primarily emphasizes enhancing accuracy but often overlooks the reliability of these models. However, the ability of a diagnostic method to maintain consistent performance under variable conditions and deliver reliable outcomes is equally vital.

Sensors perform a crucial role in monitoring and transmitting data under complex equipment operating conditions. Fault diagnosis of a device can be performed based on the measured values [4]. However, such devices can operate in extremely diverse and challenging environments, such as inertial navigation systems operating in complex environments [5]. These conditions can lead to perturbations in sensor data transmission [6]. Therefore, it is necessary to analyze and reduce the impact of these uncertainties on the sensor state when designing fault diagnostic models.

In recent years, equipment fault diagnosis considering perturbations has received much attention. Tang et al.[7] developed a model to assess the safety of complex systems under perturbative conditions. Zhang et al.[8] enhanced the diagnostic accuracy of rolling bearing performance by integrating perturbation variables into existing rules. Jena et al.[9] proposed a classification system for power network perturbations using Random Forest and Empirical Wavelet Transform techniques. Liu et al.[10] explored the resilience of digital circuits against transient disturbances through targeted circuit simulations.

There are two main challenges to this research: First, in the field of fault diagnosis, many data-driven diagnostic models face the "black box" problem, the internal working mechanism of the model is opaque, which makes it difficult to interpret and verify the diagnostic results. This opacity reduces the credibility of the diagnostic results. In addition, for many important equipment, access to high quality and large amounts of data is often difficult. With limited data samples, data-driven methods are prone to "overfitting". To solve these problems, combining expert knowledge to adjust and optimize the model is an effective way to improve the model's interpretability and reliability. Second, most research has focused on analyzing the impact of disturbances on results, often ignoring the importance of disturbance analysis in the process of reasoning or improving resistance to disturbances. In the face of unstable environmental conditions and data disturbances, it is critical to understand and analyze the performance of complex systems in the presence of disturbances. Scientific interference analysis can reveal the specific impact of interference on results and provide new strategies for system design to enhance system robustness. Using scientific methods to quantify disturbances and developing optimization strategies based on these quantitative results not only improves the system's resistance to disturbances, but also maintains performance and stability when the system is subjected to unforeseen disturbances. In general, improving the transparency and robustness of the fault diagnosis model is a key step in enhancing its reliability and effectiveness, which is important for ensuring the continuity and safety of equipment operation.

The BRB model, a semi-quantitative method, was proposed by Yang et al. [11]. The BRB method is based on fuzzy reasoning, the IF-THEN rule and the ER, which can address the uncertainty and fuzzy nature of information collection [12][13]. Because BRBs incorporate expert insights, they are more interpretable [14] and can produce reliability diagnosis results. It follows that BRB is reliability. The BRB is widely used in the field of fault diagnosis. For example, Xu et al. [15] proposed a BRB-based marine diesel engine fault diagnosis expert system. Li et al. [16] introduced an adaptive and interpretable BRB-based fault diagnosis model for complex systems. Feng et al. [17] proposed an online fault diagnosis and fault tolerance system for vehicle sensors based on a collaborative BRB involving multiple experts.

Robustness refers to the reliability of the model. Robustness plays a key role in the effectiveness of BRB modeling, effectively enhancing the interpretability of BRB results. In a study related to robustness, Cao et al. [18] proposed an analytical method for determining BRB robustness and conducted a comprehensive robustness evaluation of BRB.

To address the two problems of robustness analysis and robustness optimization of fault diagnosis models, a reliable fault diagnosis method based on BRB considering robustness analysis is proposed. First, the robustness of each part of the BRB is systematically analyzed, and the robustness constraint strategy is specified according to the sensitivity of each part's parameters. Second, the P-CMA-ES with constraints is developed according to the robustness constraint strategy to optimize the robustness and accuracy of the model.

In conclusion, the introduction of these methodologies provides a comprehensive solution that substantially improves the accuracy and reliability of fault diagnosis systems, fulfilling the essential requirement for models to adapt to and withstand operational disturbances. Traditional methods often fail to address the complex interaction between parameter sensitivity and diagnostic stability. This progress not only fills the existing gap but also provides a new benchmark for the future development and research of fault diagnosis system.

The organization of this paper is outlined as follows: Section 2 introduces the foundational concepts of the BRB and outlines the problem statement. The analysis of BRB model robustness and the application of robustness constraints are detailed in Section 3. Section 4 describes the optimization algorithm designed to operate under these robustness constraints. An empirical evaluation of the proposed fault diagnosis method is conducted in Section 5. The paper concludes with a summary in Section 6.

## 2 Basic knowledge and description of problem for BRB

### 2.1 Description of the BRB

In a BRB, the basics of *the* *k*th rule can be described as follows:

$$\begin{aligned} &\text{IF}\left(X_1 \ is\ A_1^k\right)\Lambda\left(X_2 \ is\ A_2^k\right)\Lambda\cdots\Lambda\left(X_{M_k} \ is\ A_{M_k}^k\right), \\ &\text{THEN}\left\{\left(D_1,\beta_1^k\right),\left(D_2,\beta_2^k\right),\ldots,\left(D_N,\beta_N^k\right)\right\},\ \left(\sum_{i=1}^{N}\beta_i^k \le 1\right), \\ &\text{with rule weights } \theta_k\left(k = 1,2,\ldots,L\right) \\ &\text{and attribute weights } \delta_i(i = 1,2,\ldots,M). \end{aligned} \tag{1}$$

where $X_{M_k}$ is defined as the *M*th attribute under the *k*th rule. $A_i^k\left(i=1,2,\ldots,M_k\right)$ is defined as the reference value of the *i*th prefix attribute. $\theta_k$ is defined as the weight of the kth rule. $\delta_i$ is defined as the weight of the *i*th attribute. *L* is defined as the number of rules. *M* is defined as the number of attributes. $\beta_i^k\left(i=1,2,\ldots,N\right)$ is defined as the degree of confidence of the *i*th result of the *k*th rule.

### 2.2 Reasoning of the BRB

The theoretical inference process of BRB is usually divided into five steps: input conversion, rule activation weight calculation, matching degree normalization, rule aggregation and utility calculation.

**Step 1:** Input Conversion

First, the input information is converted into a reference value distribution.

$$S(x_i)=\{(A_{i,j},a_{i,j}),i=1,2,...,M;j=1,2,...,J_m\}. \tag{2}$$

where $A_{i,j}$ is defined as the *j*th reference value of the *i*th attribute. $a_{i,j}$ is defined as the distribution level of the reference value.

**Step 2:** Matching Degree Calculation

The reference value distribution is calculated as the match degree, and the calculation process is given as follows:

$$\alpha_k=\prod_{i=1}^{T}(a_i^k)^{\bar{\delta}_i} \tag{3}$$

$$\bar{\delta}_i=\delta_i/\max_{i=1,2,...,T}\{\delta_i\}, \tag{4}$$

where $\bar{\delta}_i$ is defined as the attribute normalized weights. $\alpha_k$ is defined as the match degree of the *k*th rule.

**Step 3:** Rule Activation Weight Calculation

The rule needs to calculate whether it is activated by the activation weight, and the calculation process is given as follows:

$$w_k = \frac{\theta_k \alpha_k}{\sum_{l=1}^{L} \theta_l \alpha_k}, \tag{5}$$

where $w_k$ is defined as the weight of the activation of the rule.

**Step 4:** Rule **aggregation**

The rule aggregation process is the process of fusing all active rules to generate the final confidence distribution, and its calculation process is given as follows:

$$\beta_n = \frac{\left[\prod_{k=1}^{L}\left(w_k \beta_n^k + \gamma_{n,i}^k\right) - \prod_{k=1}^{L}\left(\gamma_{n,i}^k\right)\right]}{\left[\sum_{n=1}^{N}\prod_{k=1}^{L}\left(w_k \beta_n^k + \gamma_{n,i}^k\right) - (N-1)\prod_{k=1}^{L}\left(\gamma_{n,i}^k\right) - \prod_{k=1}^{L}(1-w_k)\right]}, \tag{6}$$

$$\gamma_{n,i}^k = 1 - w_k \sum_{i=1}^{N} \beta_i^k. \tag{7}$$

where $\beta_n$ denotes the confidence level of the result. $\gamma$ denotes an intermediate parameter.

The resulting confidence levels are distributed as follows:

$$S\left(A^*\right) = \left\{\left(D_n, \beta_n\right); n = 1,2,...,N\right\}, \tag{8}$$

where $A^*$ is defined as the vector of inputs.

**Step 5:** Utility Calculation

The final utility of $S\left(A^*\right)$ is calculated as follows:

$$u\left(S\left(A^*\right)\right) = \sum_{n=1}^{N} u\left(D_n\right)\beta_n, \tag{9}$$

where $u\left(D_n\right)$ denotes the utility value.

### 2.3 Description of the problem

The following problems still exist in the current BRB model applied for fault diagnosis.

Problem 1: How can a systematic robustness study of BRB fault diagnosis models be performed?

Within the realm of fault diagnosis methodologies, BRB models are particularly noted for their effectiveness in handling uncertain and incomplete data. However, their robustness—defined as the stability of outputs in response to parameter perturbations—remains a significant challenge, particularly in environments with noisy inputs. This study focuses on a detailed analysis centered on enhancing the robustness of BRB models, with the goal of improving their stability and reliability in real-world diagnostic applications.

Problem 2: How can the robustness of the BRB fault diagnosis model be enhanced?

Enhancing the robustness of BRB models is vital for improving their diagnostic precision in fault diagnosis scenarios. These models frequently suffer from decreased effectiveness due to their sensitivity to data perturbations, which can lead to erroneous diagnostics. This research focuses on creating and applying innovative optimization strategies that strengthen the resilience of BRB models against such disturbances, with a particular emphasis on mitigating the impact of noise and outliers. Through these strategies, the aim is to achieve a more reliable and accurate model performance under varied and unpredictable conditions.

Problem 1 and Problem 2 are addressed in Sections 3 and 4, respectively.

## 3 Robustness analysis and robustness constraints of BRB

The robustness of BRBs is critical to modeling performance and security. According to [18], robustness analysis includes the following four parts: input transformation, matching degree calculation, matching degree normalization, and rule aggregation. The Lipschitz constant is used as a criterion for robustness.

(1) There are two types of input transformation:

Type 1: $x' \le u\left(A_{j+1}\right),\ u\left(A_j\right) \le x$

Lipschitze constant is calculated as follows:

$$\vartheta_{IT-1} = \frac{\left|-a'_j + a_j + a'_{j+1} - a_{j+1}\right|}{\left|x' - x\right|} \tag{10}$$

Type 2: $u\left(A_j\right) \le x \le u\left(A_{j+1}\right), u\left(A_{j'}\right) \le x' \le u\left(A_{j'+1}\right)$

Lipschitze constant is calculated as follows:

$$\vartheta_{IT-2} = \frac{\left|-a'_{j'} + a_j + a'_{j'+1} - a_{j+1}\right|}{\left|x' - x\right|} \tag{11}$$

The input transformation part of the Lipschitz constant is calculated as follows:

$$\vartheta_{IT} = \max\left\{\vartheta_{IT-1}, \vartheta_{IT-2}\right\} \tag{12}$$

where $\vartheta_{IT}$ is the input transform of the Lipschitz constant.

**Strategy 1: The reference value should give the reference interval.**

From the above analysis, it can be seen that the reference value division has a large influence on the lipschitze constant of the input conversion part of the BRB. Fixed reference values may decrease the robustness of the model. Moreover, when experts divide the initial reference values, the reference value division may not be completely accurate due to the incompleteness of knowledge. Therefore, it is necessary to constrain the reference values to an interval. The reference interval can be given as:

$$A_i^m \in [a_{i-lb}^m, a_{i-ub}^m] \tag{13}$$

where $a_{i-lb}^m$ denotes the lower bound of the reference value given by the expert and $a_{i-ub}^m$ denotes the upper bound of the reference value given by the expert.

(2) The Lipschitz constant for the matching degree calculation is calculated as follows:

$$\vartheta_{MDC} = \max\left\{\left|\overline{\delta}_i \bullet (a_i^k)^{\overline{\delta}_i - 1} \bullet \prod_{\substack{j=1 \\ i \neq j}}^{T} (a_j^k)^{\overline{\delta}_j}\right|\right\} \tag{14}$$

where $\vartheta_{MDC}$ is the Lipschitz constant for the matching degree calculation.

**Strategy 2: Attributes should not be underweighted.**

According to the above analysis, the attribute weights have a high impact on the lipschitze constant of the matching degree calculation, and attribute weights that are too small result in a high lipschitze constant, which reduces the robustness of the model. Therefore, it is necessary to constrain the attribute weights, and the constraint strategy can be given as:

$$\overline{\delta}_i = \begin{cases} \overline{\delta}_i, & if\ \overline{\delta}_i \geq \gamma \\ \gamma, & if\ \overline{\delta}_i < \gamma \end{cases} \tag{15}$$

where $\gamma$ is the constraint threshold given by experts in the industry.

(3) For matching degree normalization, it can be divided into two cases. When $i = k$, $k$ is defined as the number of activation rules, lipschitze constant is calculated as follows:

$$g_{i=k} = \left|\frac{\theta_k(\theta_1\alpha_1 + \theta_2\alpha_2 + \theta_3\alpha_3 + \theta_4(1 - \alpha_1 - \alpha_2 - \alpha_3 + \alpha_k) - \alpha_k\theta_k)}{((\theta_1 - \theta_4)\alpha_1 + (\theta_2 - \theta_4)\alpha_2 + (\theta_3 - \theta_4)\alpha_3 + \theta_4))^2}\right| \tag{16}$$

where $k$ is defined as the index of the activation rule and $i$ is defined as the index of the currently activated rule.

When $i \neq k$, the lipschitz constant is calculated as follows:

$$g_{i \neq k} = \left|\frac{-\theta_k\alpha_k(\theta_i - \theta_4)}{((\theta_1 - \theta_4)\alpha_1 + (\theta_2 - \theta_4)\alpha_2 + (\theta_3 - \theta_4)\alpha_3 + \theta_4))^2}\right| \tag{17}$$

where $g_{i=k}$ is the Lipschitz constant for $i = k$ and $g_{i \neq k}$ is the Lipschitz constant for $i \neq k$.

$$\vartheta_{MDN} = \max\left\{g_{i=k}, g_{i \neq k}\right\} \tag{18}$$

where $\vartheta_{MDN}$ is the Lipschitz constant of the matching degree normalization.

**Strategy 3: The weights of the rules should not be underweighted.**

According to the above analysis, the rule weights have a high impact on the lipschitze constant of the match normalization; moreover, lower rule weights will cause redundancy in the structure of the BRB model, and both the interpretability and robustness of the BRB model will be limited. Therefore, it is necessary to constrain the rule weights, and its constraint strategy is given as:

$$\theta_i = \begin{cases} \theta_i, & if\ \theta_i \geq \tau \\ \tau, & if\ \theta_i < \tau \end{cases} \tag{19}$$

where $\tau$ is the constraint threshold given by experts in the industry.

(4) The Lipschitz constant of rule aggregation is calculated as follows:

$$\vartheta_{ER} = \max\{\phi_{n,k}, \rho_{\beta_{n,k}}, \rho_{\beta_{\bar{n},k}}\} \tag{20}$$

where $\vartheta_{ER}$ is the Lipschitz constant of the rule aggregation.

When $n \neq \tilde{n}$, $\rho_{\beta_{n,k}}$ represents the Lipschitz constant of the effect of $\beta_{n,k}$ on $\partial\hat{\beta}_{\tilde{n}}$. When $n = \tilde{n}$, $\rho_{\beta_{\tilde{n},k}}$ represents the Lipschitz constant of the effect of $\beta_{\tilde{n},k}$ on $\partial\hat{\beta}_{\tilde{n}}$.

when $n \neq \tilde{n}$

$$\rho_{\beta_{n,k}} = \frac{-\left[\prod_{i=1}^{4}(w_i(\beta_{n,i}-1)+1)-\prod_{i=1}^{4}(1-w_i)\right]\bullet\left[w_k\prod_{i\neq k}^{4}(w_i(\beta_{n,i}-1)+1)\right]}{\left[\sum_{n=1}^{N}\prod_{i=1}^{4}(w_i(\beta_{n,i}-1)+1)-N\prod_{i=1}^{4}(1-w_i)\right]^2} \tag{21}$$

where $n$ is defined as the index of the confidence level of the activation rule and $\tilde{n}$ is defined as the index of the confidence level of the currently activated rule.

when $n = \tilde{n}$

$$\rho_{\beta_{\tilde{n},k}} = \frac{w_k\prod_{i\neq k}^{4}(w_i(\beta_{\tilde{n},i}-1)+1)\bullet\left[\sum_{n=1}^{N}\prod_{i=1}^{4}(w_i(\beta_{n,i}-1)+1)-N\prod_{i=1}^{4}(1-w_i)\right]-\left[\prod_{i=1}^{4}(w_i(\beta_{\tilde{n},i}-1)+1)-\prod_{i=1}^{4}(1-w_i)\right]\bullet\left[w_k\prod_{i\neq k}^{4}(w_i(\beta_{\tilde{n},i}-1)+1)\right]}{\left[\sum_{n=1}^{N}\prod_{i=1}^{4}(w_i(\beta_{n,i}-1)+1)-N\prod_{i=1}^{4}(1-w_i)\right]^2} \tag{22}$$

$$\phi_{n,k} = \frac{\left[(\beta_{n,k}-1)\prod_{i\neq k}^{4}(w_i(\beta_{n,i}-1)+1)+\prod_{i\neq k}^{4}(1-w_i)\right]\bullet\left[\sum_{n=1}^{N}\prod_{i=1}^{4}(w_i(\beta_{n,i}-1)+1)-N\prod_{i=1}^{4}(1-w_i)\right]-\left[\prod_{i=1}^{4}(w_i(\beta_{n,i}-1)+1)-\prod_{i=1}^{4}(1-w_i)\right]\bullet\left[\sum_{n=1}^{N}(\beta_{n,k}-1)\prod_{i\neq k}^{4}(w_i(\beta_{n,i}-1)+1)+N\prod_{i\neq k}^{4}(1-w_i)\right]}{\left[\sum_{n=1}^{N}\prod_{i=1}^{4}(w_i(\beta_{n,i}-1)+1)-N\prod_{i=1}^{4}(1-w_i)\right]^2} \tag{23}$$

where $\phi_{n,k}$ represents the Lipschitz constant of the effect of $w_k$ on $\hat{\beta}_n$.

(5) The Lipschitz constant of the BRB is calculated as follows:

$$\vartheta_{BRB} = \vartheta_{IT}\bullet\vartheta_{MDC}\bullet\vartheta_{MDN}\bullet\vartheta_{ER} \tag{24}$$

where $\vartheta_{BRB}$ is the Lipschitz constant of the BRB.

According to the above score, the robustness of BRB is significantly affected by its inputs and parameters. Therefore, precisely controlling and limiting the key parameters of BRB is necessary to improve its robustness. In engineering practice, by enhancing the robustness of BRBs, the risk of BRBs in the face of uncertainties can be effectively reduced, thus improving the overall safety and reliability.

## 4 Optimization methods considering robustness constraints

### 4.1 Objective function with robustness constraints

The objective function, incorporating robustness constraints, is defined by the equation below:

$$
\begin{aligned}
&\min \Phi(A,\theta,\delta,\beta) \\
&s.t.\ 0 \le \theta_k \le 1,\ \ 0 \le \delta_m \le 1,\ \ 0 \le \beta_{n,k} \le 1, \\
&\quad \sum\nolimits_{n=1}^{N} \beta_{n,k} \le 1,\ \ a_{i-lb}^m \le A_i^m \le a_{i-ub}^m, \\
&\quad \bar{\delta}_i = \begin{cases} \bar{\delta}_i, & if\ \bar{\delta}_i \ge \gamma \\ \gamma, & if\ \bar{\delta}_i < \gamma \end{cases}, \theta_i = \begin{cases} \theta_i, & if\ \theta_i \ge \tau \\ \tau, & if\ \theta_i < \tau \end{cases}, \\
&\quad (k = 1,\ldots,L,\ m = 1,\ldots,M,\ n = 1,\ldots,N,\ i = 1,\ldots,T)
\end{aligned} \tag{25}
$$

where $\Phi(A,\theta,\delta,\beta)$ represents the discrepancy between the output of the BRB and the actual observed value.

### 4.2 A modified P-CMA-ES algorithm

This study explores the robustness constraints in the model optimization process. In BRB-based fault diagnosis models, the accuracy and robustness of the BRB are affected not only by the reference point but also by variations in other parameters [19]. Therefore, the focus of this paper in the model optimization process is to refine the reference value, attribute weights, and rule weights with the aim of improving accuracy while ensuring robustness.

In the present stage of research, many optimization algorithms are used as optimization models. A notable optimization algorithm frequently applied to the BRB model is the P-CMA-ES. This method is recognized for its effectiveness in handling complex optimization problems, particularly in adjusting parameters within BRB models to achieve enhanced accuracy and efficiency[20]. It is suitable for addressing complex optimization issues that are nonlinear and nonconvex in continuous settings. This study employs the P-CMA-ES to fine-tune the parameters of the model created by D-BRB, aiming to further enhance the accuracy of the model.

The P-CMA-ES unfolds through the following steps:

**Step 1:** Determine the initial parameters to be optimized $w^0 = \Omega^0$ and the initial parameters of the P-CMA-ES.

$$
\Omega^0 = \left\{ \theta_1,\ldots,\theta_L,\beta_{1,1},\ldots,\beta_{L,N},\delta_1,\ldots,\delta_M \right\} \tag{26}
$$

where $\Omega^0$ is defined as the initial vector of parameters subject to optimization and $w^0$ is defined as the initial mean value.

**Step 2:** Determine the objective function and the constraints. The MSE quantifies the modeling accuracy of the BRB and is calculated using the following formula:

$$
MSE\left(\theta_k,\beta_{n,k},\delta_i\right) = \frac{1}{T}\sum\nolimits_{t=1}^{T}(result_{autual} - result_{predict})^2 \tag{27}
$$

where $T$ represents the total number of observational data points, $result_{autual}$ is the actual output of the system, and $result_{predict}$ is the diagnostic result of the BRB.

Based on the above definition, the objective function and its associated constraints are specified as follows:

$$\begin{aligned}
&\min MSE\left(\theta_k, \beta_{n,k}, \delta_i\right) \\
&s.t.\ 0 \le \theta_k \le 1,\ 0 \le \beta_{n,k} \le 1, \\
&0 \le \delta_i \le 1,\ \sum_{n=1}^{N} \beta_{n,k} \le 1 \\
&k = 1,2,\ldots,L,\ n = 1,2,\ldots,N,\ i = 1,2,\ldots,M
\end{aligned} \tag{28}$$

**Step 3:** Perform a sampling operation to generate the initial population:

$$\Omega_i^{g+1} \sim w^g + \varepsilon^g \mathbb{N}(0, C^g) \qquad i = 1,\ldots,\lambda \tag{29}$$

where $\Omega_i^{g+1}$ represent the *i*th solution in the *(g+1)*th generation. $\varepsilon$ represent the step size, and $\mathbb{N}$ represent the normal distribution. $C^g$ represent the covariance matrix in the *g*th generation.

**Step 4:** Perform the projection operations to ensure adherence to the constraints:

$$\begin{aligned}
\Omega_i^{g+1}(1+n_e \times (j-1) : n_e \times j) = \Omega_i^{g+1}(1+n_e \times (j-1) : n_e \times j) - A_e^T \times (A_e \times A_e^T)^{-1} \\
\times \Omega_i^{g+1}(1+n_e \times (j-1) : n_e \times j) \times A_e
\end{aligned} \tag{30}$$

where hyperplane can be denoted by $A_e \Omega_i^g (1+n_e \times (j-1) : n_e \times j) = 1$, $n_e$ represents the count of variables involved in the equality constraint of the solution $\Omega_i^g$ , $j = 1,\ldots,N+1$ indicates the quantity of equality constraints present in the solution $\Omega_i^g$ , and $A_e = [1 \cdots 1]_{1 \times N}$ represent the parameter vector.

**Step 5:** Perform selection to revise the mean based on $w^{g+1} = \sum_{i=1}^{\tau} h_i \Omega_{i:\lambda}^{g+1}$ , where $h_i$ denotes the weight coefficient of the ith solution. $\Omega_{i:\lambda}^{g+1}$ represent the *i*th solution from $\lambda$ solutions in the *(g+1)*th generation. $\tau$ represent the size of the offspring population.

**Step 6:** Carry out adaptation to refine the covariance matrix.

$$C^{g+1} = (1-c_1-c_2) \bullet C^g + c_1 p_c^{g+1} (p_c^{g+1})^T + c_2 \sum_{i=1}^{\tau} h_i \left(\frac{\Omega_{i:\lambda}^{g+1} - w^g}{\varepsilon^g}\right)\left(\frac{\Omega_{i:\lambda}^{g+1} - w^g}{\varepsilon^g}\right)^T \tag{31}$$

$$p_c^{g+1} = (1-c_c) \bullet p_c^g + \sqrt{c_c(2-c_c)} \bullet \left(\sum_{i=1}^{\tau} h_i^2\right)^{-0.5} \bullet \frac{(w^{g+1} - w^g)}{\varepsilon^g} \tag{32}$$

where the step size is modified based on the subsequent equation:

$$\varepsilon^{g+1} = \varepsilon^g \exp\left(\frac{c_\sigma}{d_\sigma}\left(\frac{\left\| p_\sigma^{g+1} \right\|}{E\left\| N(0,1) \right\|} - 1\right)\right) \tag{33}$$

$$p_\sigma^{g+1} = (1-c_c) \bullet p_\sigma^g + \sqrt{c_c(2-c_c)} \bullet \left(\sum_{i=1}^{\tau} h_i^2\right)^{-0.5} \bullet \frac{(w^{g+1} - w^g)}{\varepsilon^g} \bullet (C^g)^{-0.5} \tag{34}$$

where $c_1$ and $c_2$ are defined as the learning rates, $p_c$ is defined as the evolution path, and $c_c$ is identified as the retrospective duration of the evolutionary path.

**Step 7:** Repeat the process iteratively until the optimal solution is reached $\Omega_{optimal}$ . The optimal BRB is then modeled.

## 5 Case study

### 5.1 Fault diagnosis of the WD615 diesel engine

#### 5.1.1 Background description

Diesel engines are used in a wide variety of applications including machinery, electricity, and other fields. Consequently, it is crucial to perform effective troubleshooting on diesel engines. Given their complexity and significance, accurate and dependable models are essential for diagnosing these issues. Hence, it is necessary to develop a robust BRB model for fault diagnosis. In this research, the WD615 diesel engine serves as the case study. The engine operated at a steady-state speed of 1800 rpm, and the data was collected at a sampling frequency of 12.8 kHz, and the mean and kurtosis were used as prerequisite attributes[21]. The distribution of the data is depicted in Fig.1

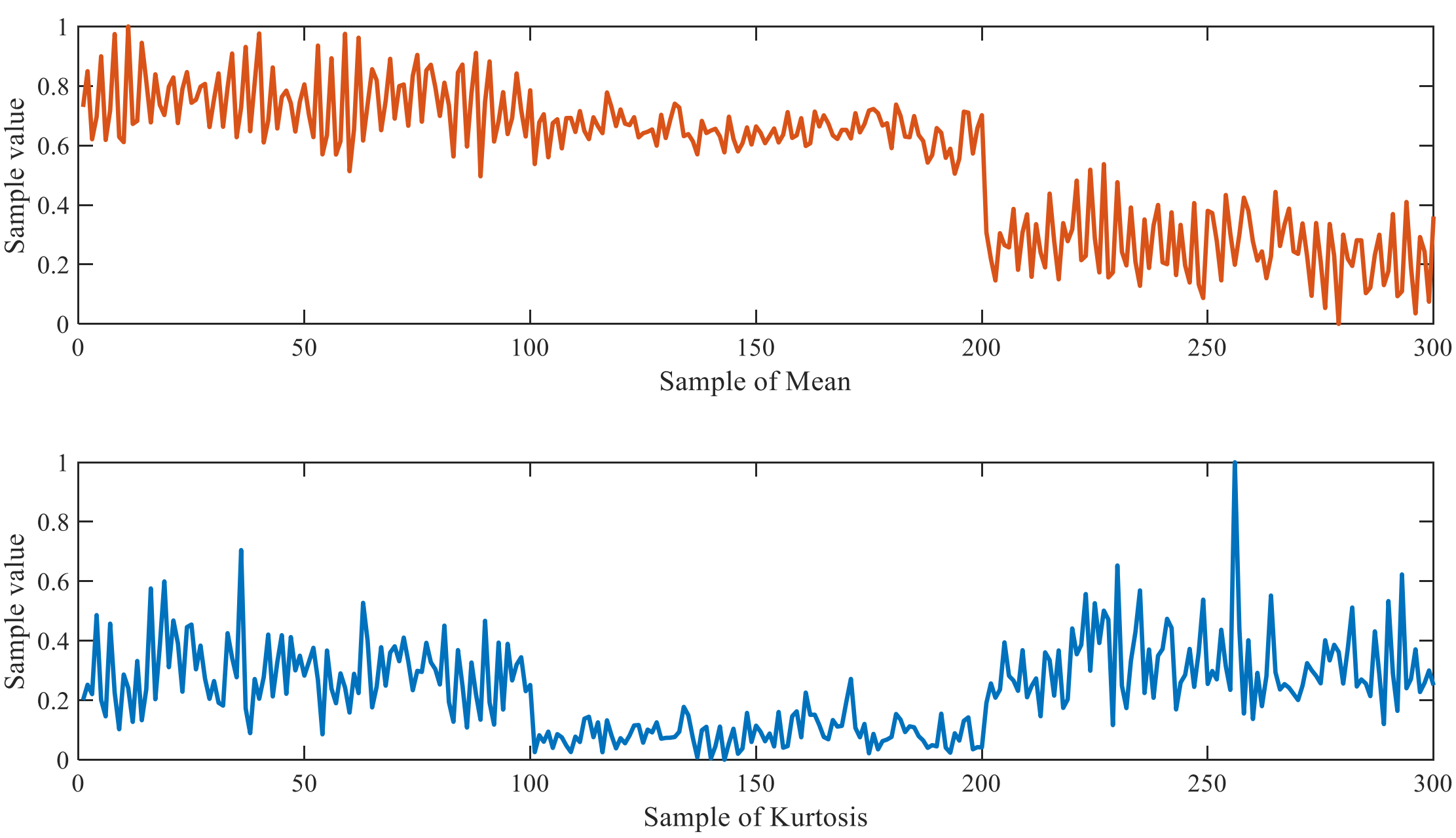

Fig. 1: Data distribution of the means and kurtosis

#### 5.1.2 Initial BRB construction

This paper extracts a total of 300 datasets, which are divided into two parts: 150 for training and 150 for testing. The initial parameters of the BRB, as detailed in the table, are supplied by domain experts who bring profound knowledge and insights from the relevant fields, contributing valuable and reliable inputs for developing the BRB. Their expertise underpins the construction of BRBs, guaranteeing their precision and efficacy in modeling complex decision-making processes. The initial parameters of the constrained P-CMA-ES are shown in Table 1. The reference values and constraints for the mean and kurtosis are shown in Table 2. Failure status in three categories, normal (N), moderate failure (m), and severe failure (S), is shown in Table 3. The initial belief distribution is outlined in Table 4. The kth rule within the BRB fault diagnosis model is formulated as follows:

$$IF\left(\text{Mean is } A_{Mean}^{k}\right)\wedge\left(\text{Kurtosis is } A_{Kurtosis}^{k}\right)$$
$$\text{THEN fault state is}\left\{\left(N,\beta_{N}^{k}\right),\left(M,\beta_{M}^{k}\right),\left(S,\beta_{S}^{k}\right)\right\},$$
$$\text{with } \theta_{k}\left(k=1,2,\ldots,16\right) \text{ and } \delta_{i}\left(i=1,2\right)$$

where $A_{Mean}^{k}$ and $A_{Kurtosis}^{k}$ represents the reference intervals of the mean and kurtosis, respectively. $\beta_{N}^{k}$ $\beta_{M}^{k}$ and $\beta_{S}^{k}$ represents the belief degrees for fault states N, M, and S, respectively.

Table 1: Initial settings for the P-CMA-ES

| | |
|---|---|
| step leader | 0.5 |
| Number of iterations | 200 |
| Attribute weight constraint $\gamma$ | 0.8 |
| Rule weight constraint $\tau$ | 0.8 |

Table 2: Reference intervals for the mean and kurtosis

| Attribute | Attribute weight | L | BM | M | H |
|---|---|---|---|---|---|
| Mean | 0.8 | [-0.2-0.0] | [0.23-0.45] | [0.60-0.75] | [1.0-1.2] |
| Kurtosis | 0.8 | [-0.2-0.0] | [0.11-0.20] | [0.35-0.60] | [1.0-1.2] |

Table 3: The reference values for the results

| | N | M | S |
|---|---|---|---|
| Result | 1 | 0.5 | 0 |

Table 4: The initial belief distribution for the BRB

| NO. | Attribute | | The rule weight | The initial belief |
|---|---|---|---|---|
| | Mean | Kurtosis | | {β1, β2, β3} |
| 1 | L | L | 0.80 | {0.00,0.00,1.00} |
| 2 | L | BM | 0.80 | {0.00,0.20,0.80} |
| 3 | L | M | 0.80 | {0.20,0.40,0.40} |
| 4 | L | H | 0.80 | {0.20,0.30,0.50} |
| 5 | BM | L | 0.80 | {0.35,0.35,0.30} |
| 6 | BM | BM | 0.80 | {0.60,0.30,0.10} |
| 7 | BM | M | 0.80 | {0.80,0.15,0.05} |
| 8 | BM | H | 0.80 | {0.80,0.10,0.10} |
| 9 | M | L | 0.80 | {0.20,0.60,0.20} |
| 10 | M | BM | 0.80 | {0.30,0.30,0.40} |
| 11 | M | M | 0.80 | {0.10,0.10,0.80} |
| 12 | M | H | 0.80 | {0.20,0.40,0.40} |
| 13 | VS | L | 0.80 | {0.10,0.10,0.80} |
| 14 | H | BM | 0.80 | {0.00,0.10,0.90} |
| 15 | H | M | 0.80 | {0.20,0.20,0.60} |
| 16 | H | H | 0.80 | {0.30,0.30,0.40} |

### 5.1.3 Optimization of BRB

To enhance the accuracy and robustness of the model, the optimization process employs the constrained P-CMA-ES. The refined parameters following this optimization are detailed in Tables 5 and 6. Throughout this paper, the classic BRB model is named to as BRB0, while the enhanced model developed in this study is designated BRB1. Fig.2 depicts the weights of optimized rules, highlighting their significance in the system. The red line shows the expert-set optimization threshold, and the optimization is conducted within this specified range. As depicted in Fig.3, the diagnostic performance of BRB1 is shown to superior that of BRB0, indicating significant improvements in accurate by the applied optimization strategies.

Table 5: The optimized reference values for attributes

| Attribute | Attribute weight | L | BM | M | H |
|---|---|---|---|---|---|
| Mean | 0.9405 | -0.1975 | 0.45 | 0.6269 | 1.0001 |
| Kurtosis | 0.9441 | -0.0001 | 0.1895 | 0.3528 | 1.889 |

Table 6: The optimized belief distribution of the BRB

| NO. | Attribute | | The rule weight | The initial belief |
|---|---|---|---|---|
| | Mean | Kurtosis | | {β1, β2, β3} |
| 1 | L | L | 0.91 | {0.26,0.36,0.38} |
| 2 | L | BM | 0.89 | {0.80,0.20,0.00} |
| 3 | L | M | 0.80 | {0.82,0.18,0.00} |
| 4 | L | H | 0.99 | {0.38,0.26,0.36} |
| 5 | BM | L | 0.81 | {0.52,0.33,0.15} |
| 6 | BM | BM | 1.00 | {1.00,0.00,0.00} |
| 7 | BM | M | 1.00 | {1.00,0.00,0.00} |
| 8 | BM | H | 0.80 | {0.62,0.00,0.38} |
| 9 | M | L | 1.00 | {0.11,0.84,0.05} |
| 10 | M | BM | 0.80 | {0.10,0.31,0.59} |
| 11 | M | M | 0.80 | {0.00,0.00,1.00} |
| 12 | M | H | 0.80 | {0.35,0.03,0.62} |
| 13 | VS | L | 0.99 | {0.51,0.38,0.11} |
| 14 | H | BM | 0.94 | {0.00,0.00,1.00} |
| 15 | H | M | 0.80 | {0.00,0.00,1.00} |
| 16 | H | H | 0.81 | {0.00,0.00,1.00} |

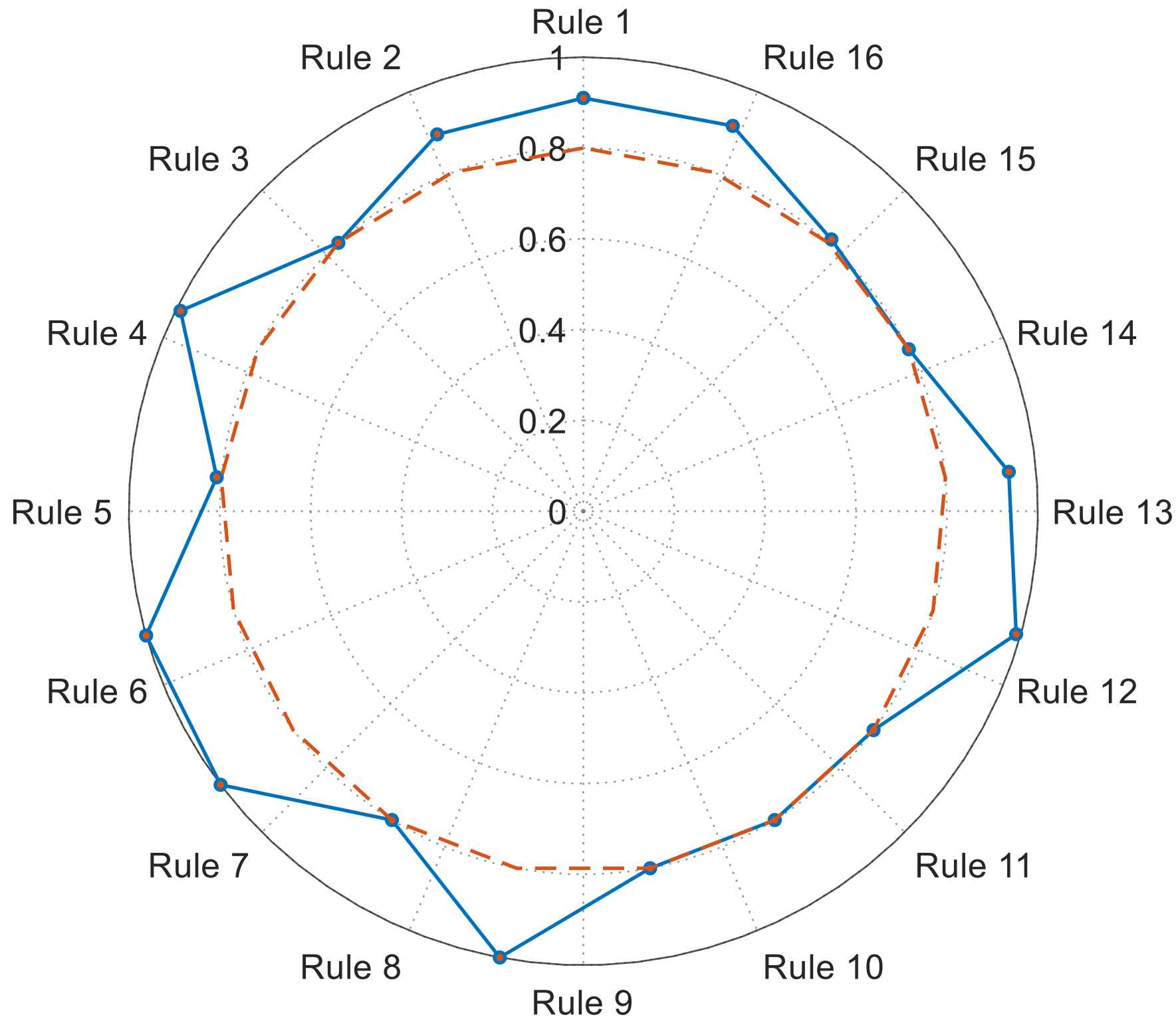

Fig.2: The optimized rule weights

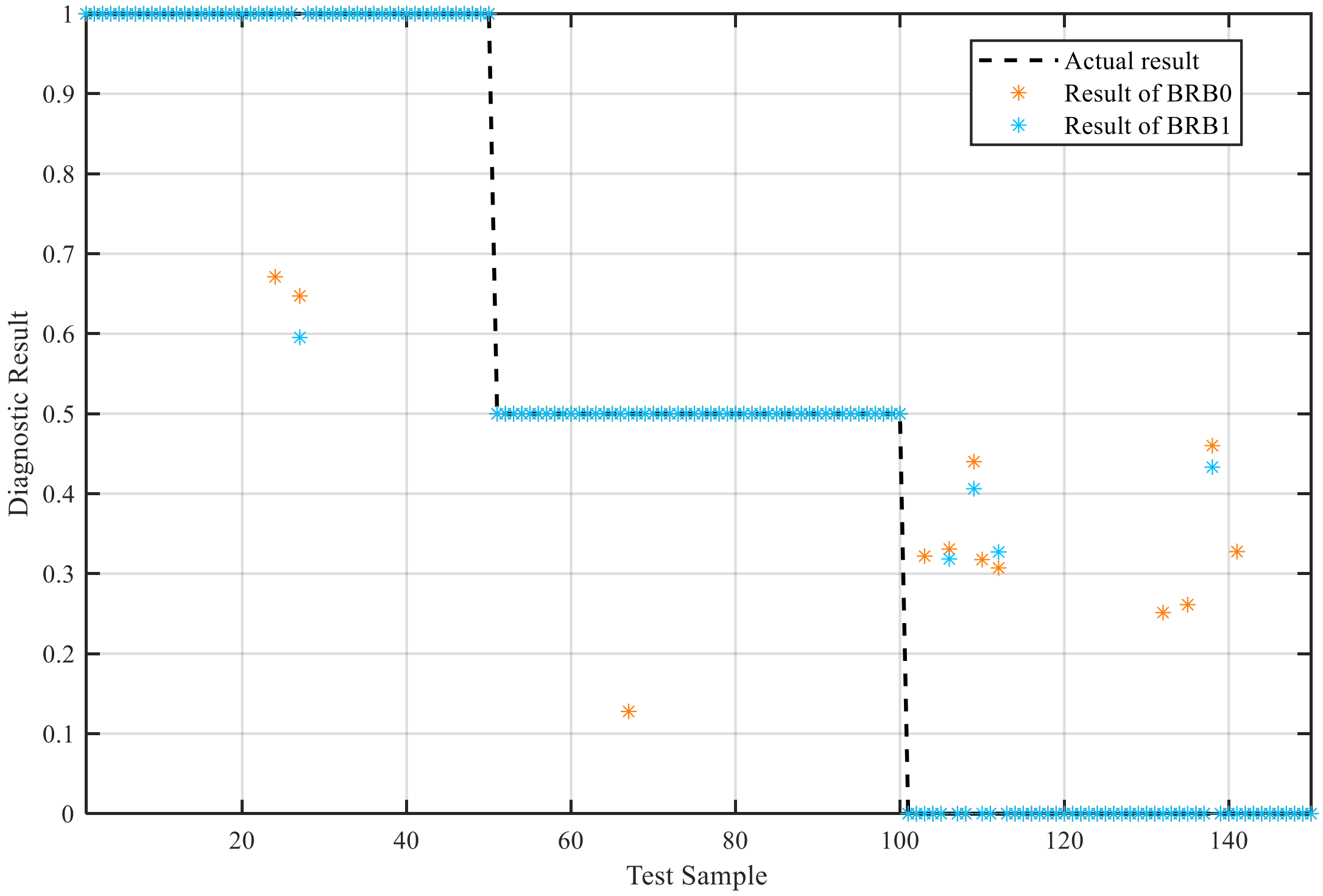

Fig. 3: Diagnostic results for BRB0 and BRB1

Table 5, along with Fig. 4 and Fig. 5, provide a clear visual representation of how the constraint strategy positively influences the optimization process of the model. Fig.4 has shown that as the number of iterations within the P-CMA-ES increases, there is a noticeable trend towards convergence, indicating effective optimization over time. The attribute weights and rule weights for both the BRB0, and the BRB1, display a pattern of stabilization through the iterative process. The data suggests that under the influence of the constraint strategy, the attributes and rules in BRB1 assume a higher level of significance compared to those in BRB0. This enhancement is indicative of the constraint strategy's crucial role in finely tuning the weights of attributes and rules within the BRB models, leading to a more robust and accurate fault diagnosis capability. These outcomes underscore the value of the constraint strategy in refining the model's parameters to achieve superior diagnostic performance.

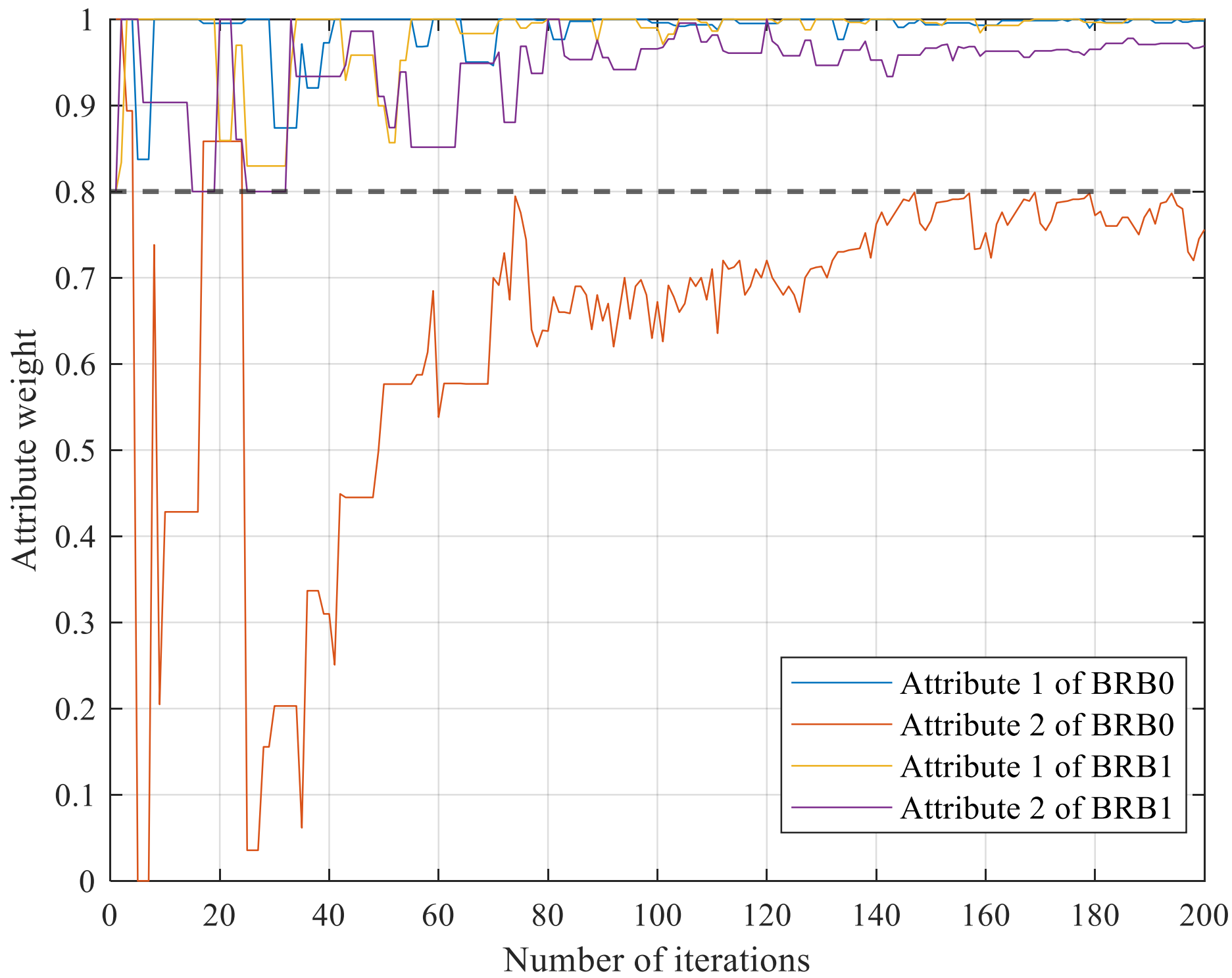

Fig. 4: Attribute weights for BRB0 and BRB1

Table 7 presents the Lipschitz constants for each component of the BRB models, BRB0 and BRB1. The data clearly indicates that the overall Lipschitz constant for BRB1 is substantially reduced compared to BRB0, with a significant improvement of 66.3% attributed to the implementation of the constraint strategy. This decrease is not only evident in the global measure but also consistently observed across various model components such as input transformation, match calculation, and match normalization in BRB1. These lower Lipschitz constants in BRB1 compared to those in BRB0 affirm the successful application of the constraint strategy, which not only enhances model stability and reduces sensitivity to input variations but also boosts the interpretability of the model. This enhancement in interpretability is crucial as it allows for easier understanding and analysis of the model's functioning, further supporting its practical application in fault diagnosis scenarios. The results demonstrate that the constraint strategy not only optimizes performance metrics but also significantly contributes to making the model more understandable and reliable.

Table 7: The Lipschitze constant of the BRB

| | $\vartheta_{IT}$ | $\vartheta_{MDC}$ | $\vartheta_{MDN}$ | $\vartheta_{ER}$ | $\vartheta_{BRB}$ |
|---|---|---|---|---|---|
| BRB0 | 14.2755 | 1.1922 | 2.7161 | 1.03 | 47.6135 |
| BRB1 | 12.2942 | 1.0371 | 1.2265 | 1.03 | 16.0478 |

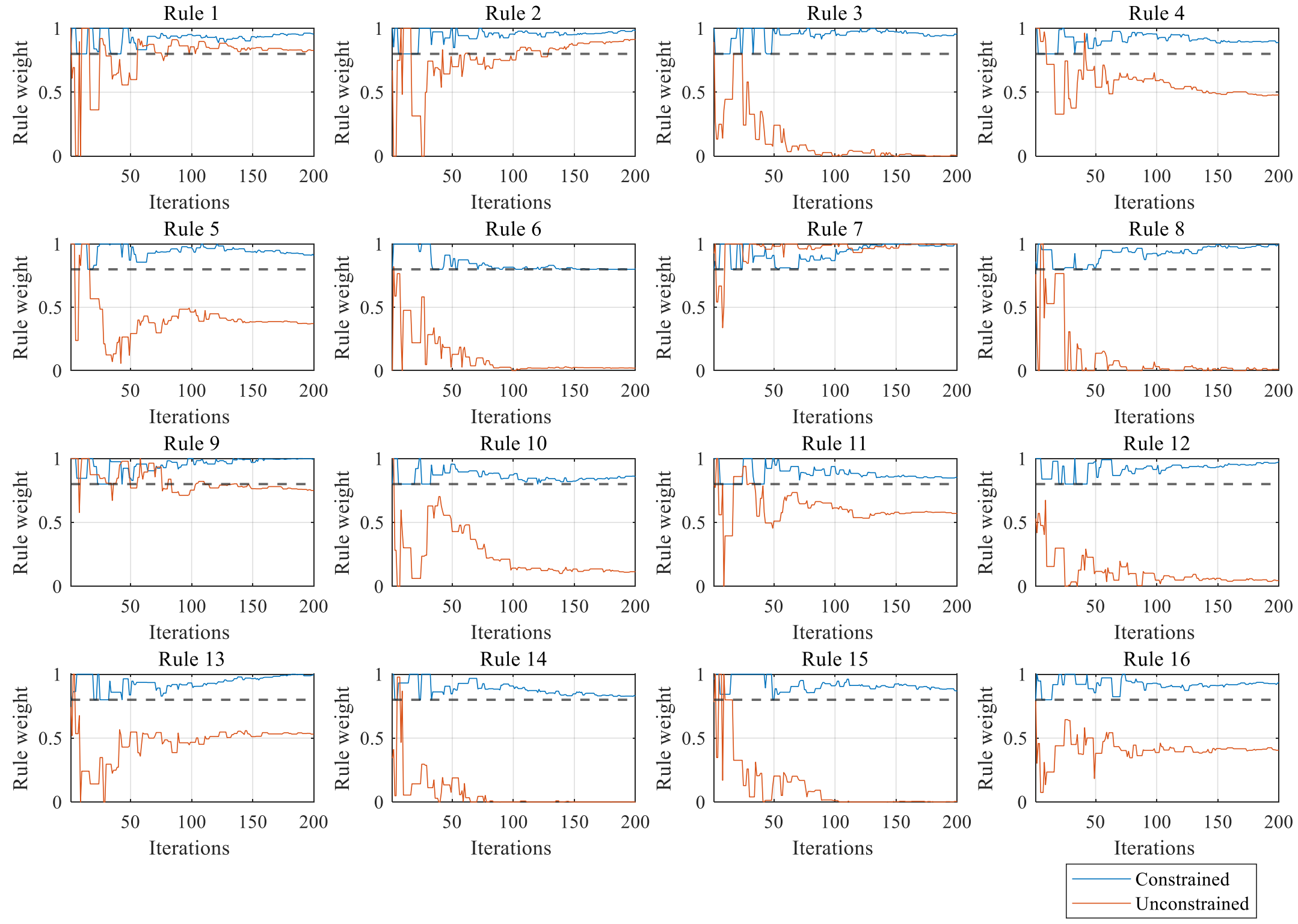

Fig. 5: Rule weights for BRB0 and BRB1

### 5.1.4 Comparative study

To further investigate the superior performance of the proposed model, a comparative analysis is conducted using different models such as long short-term memory (LSTM) network, back propagation neural network (BPNN), and radial basis function (RBF). Accuracy, macro accuracy, macro recall, macro-F1-score, micro accuracy, micro recall and micro-F1-score were used as evaluation metrics.

The data presented in Table 8 meticulously delineates a comparative analysis of the BRB model, specifically the BRB1 , against its contemporaries in the realm of fault diagnosis. Notably, the BRB1 model demonstrates an unequivocal superiority across all metrics when juxtaposed with the BRB0 model. This superiority is conspicuously manifest in the macro-F1 score, a metric that encapsulates the harmonic mean of precision and recall, thereby indicating a commendable equilibrium between these critical dimensions. When positioned against neural network paradigms, such as Long Short-Term Memory (LSTM) networks and Backpropagation Neural Networks (BPNN), the BRB1 model not only achieves accuracy rate of 96.67% but also transcends these models in both macroscopic and microscopic evaluations of precision and recall. This exemplifies the BRB1 model's exceptional proficiency in accurately discerning and categorizing faults across diverse categories (macro measures), as well as its overall effectiveness in classification tasks (micro measures). Furthermore, while the Radial Basis Function (RBF) network surpasses the performance of the BPNN, it does not approximate the advanced capabilities of the BRB1 model. The latter achieves an impressive enhancement in accuracy, exceeding a 10% increment, which is particularly significant in domains where precision is of utmost importance.

Beyond mere numerical superiority, the BRB1 model's rule-based architecture endows it with a level of interpretability that is notably absent in the more opaque 'black box' models such as typical neural networks. The transparency with which the BRB1 model processes and evaluates information offers a discernible advantage in environments where understanding the rationale behind decision-making is as critical as the outcomes of those decisions. This feature is particularly vital in intricate engineering systems where diagnostic decisions necessitate full comprehension and justification.

In summary, the robust performance coupled with the high degree of interpretability of the BRB1 model distinctly positions it as a superior tool in the field of fault diagnosis, offering both high reliability and crucial insights into the decision-making processes within equipment.

Table 8: Comparison of accuracy between BRB and other models

| | Accuracy | Macro Precision | Macro Recall | Macro F1 Score | Micro Precision | Micro Recall | Micro F1 Score |
|---|---|---|---|---|---|---|---|
| BRB0 | 92.00% | 93.10% | 92.00% | 92.54% | 92.00% | 92.00% | 92.00% |
| BRB1 | 96.67% | 96.97% | 96.67% | 96.82% | 96.67% | 96.67% | 96.67% |
| LSTM | 90.67% | 92.00% | 90.67% | 91.33% | 90.67% | 90.67% | 90.67% |
| BPNN | 87.33% | 87.60% | 87.33% | 87.47% | 87.33% | 87.33% | 87.33% |
| RBF | 86.00% | 90.14% | 86.00% | 88.02% | 86.00% | 86.00% | 86.00% |

## 5.2 Generalizability study

To assess the generalization capability of the proposed model, a fault diagnosis model is developed in this section with a publicly available bearing dataset from Case Western Reserve University. The link to the dataset is as follows: https://engineering.case.edu/bearingdatacenter/welcome. Drive-side accelerometer data were selected for this study. The data were selected for a diameter range of 0.007 inches and a motor speed of 1797 RPM. There are four types of bearing failure states, namely, normal failure (N), inner ring failure (I), rolling shaft failure (R), and outer ring failure (O), as shown in Table 11. Each state contains 120 samples, and each sample has 2048 fault points. The time-domain features of the samples are extracted, and the root mean square and variance are used as prerequisite attributes. The reference values and constraints for the root mean square(RMS) and variance are shown in Table 9 and Table 10. The initial confidence distribution is described in Table 12.The *k*th rule within the BRB fault diagnosis framework is formulated as follows:

$$\begin{aligned}&IF\left(\text{RMS is } A_{RMS}^{k}\right)\wedge\left(\text{Variance is } A_{Variance}^{k}\right)\\&\text{THEN fault state is}\left\{\left(N,\beta_N^k\right),\left(I,\beta_I^k\right),\left(R,\beta_R^k\right),\left(O,\beta_O^k\right)\right\},\\&\text{with } \theta_k\left(k=1,2,\ldots,12\right)\ and\ \delta_i\left(i=1,2\right)\end{aligned}$$

where $A_{RMS}^{k}$ and $A_{Variance}^{k}$ represents the reference intervals of the RMS and variance, respectively. $\beta_N^k$ $\beta_I^k$ $\beta_R^k$ and $\beta_O^k$ represents the belief degrees for the fault states N, I, R and O, respectively.

Table 9: Reference intervals for the RMS

| Attribute | Attribute weight | L | M | H |
|---|---|---|---|---|
| RMS | 0.8 | [-0.20-0.00] | [0.30-0.60] | [1.00-1.20] |

Table 10: Reference intervals for the variance

| Attribute | Attribute weight | L | BM | M | H |
|---|---|---|---|---|---|
| Variance | 0.8 | [-0.20-0.00] | [0.33-0.37] | [0.65-0.68] | [1.00-1.20] |

Table 11: The reference values of the results

| | N | I | R | O |
|---|---|---|---|---|
| Result | 0 | 0.33 | 0.66 | 1 |

Table 12: The initial belief distribution of the BRB

| NO. | Attribute | | The rule weight | The initial belief |
|---|---|---|---|---|
| | Mean | Kurtosis | | {β1, β2, β3, β4} |
| 1 | L | L | 0.80 | {1,0,0,0} |
| 2 | L | BM | 0.80 | {0,0,0,1} |
| 3 | L | M | 0.80 | {0.1,0.2,0.3,0.4} |
| 4 | L | H | 0.80 | {0.5,0.5,0,0} |
| 5 | M | L | 0.80 | {0,0.1,0.2,0.7} |
| 6 | M | BM | 0.80 | {1,0,0,0} |
| 7 | M | M | 0.80 | {0,0.25,0.25,0.5} |
| 8 | M | H | 0.80 | {0.2,0.2,0.5,0.1} |
| 9 | H | L | 0.80 | {0,0,0.5,0.5} |
| 10 | H | BM | 0.80 | {0.5,0.2,0.15,0.15} |
| 11 | H | M | 0.80 | {0,0,0,1} |
| 12 | H | H | 0.80 | {0,0,0,1} |

The comparative analysis detailed in Tables 13 and 14 substantiates the BRB1 model's superior performance metrics, particularly when juxtaposed against the BRB0 model and alternative computational models utilized in fault diagnosis. This superiority is quantified not only in terms of the model's robustness—where the Lipschitz constants serve as a testament to its stability against input perturbations—but also in accuracy, a pivotal benchmark in fault diagnosis. The generalizability of the BRB1 model is convincingly verified by its sustained performance across various datasets, an indicator of its robustness and applicability to different fault diagnosis contexts. Such generalizability is further enhanced by the model's rule-based architecture, which affords a level of interpretability often absent in models characterized by 'black box' operations. This interpretability is not just theoretically advantageous but pragmatically vital in complex systems where the reasoning behind diagnostic decisions must be both transparent and justifiable.

Table 13: The Lipschitze constant of the BRB

| | $\vartheta_{IT}$ | $\vartheta_{MDC}$ | $\vartheta_{MDN}$ | $\vartheta_{ER}$ | $\vartheta_{BRB}$ |
|---|---|---|---|---|---|
| BRB1 | 6.0606 | 1.3563 | 7.0878 | 1.03 | 60.0058 |
| BRB2 | 6.5983 | 1.0121 | 6.4412 | 1.03 | 44.3054 |

Table 14: Comparison of accuracy between BRB and other models

| | Accuracy | Macro Precision | Macro Recall | Macro F1 Score | Micro Precision | Micro Recall | Micro F1 Score |
|---|---|---|---|---|---|---|---|
| BRB1 | 96.86% | 96.96% | 96.88% | 96.92% | 96.88% | 96.88% | 96.88% |
| BRB2 | 98.96% | 99.00% | 98.96% | 98.98% | 98.96% | 98.96% | 98.96% |
| LSTM | 81.51% | 89.37% | 81.51% | 85.26% | 81.51% | 81.51% | 81.51% |
| BPNN | 89.58% | 92.11% | 89.58% | 90.83% | 89.58% | 89.58% | 89.58% |
| RBF | 87.50% | 87.96% | 87.5% | 87.73% | 87.50% | 87.50% | 87.50% |

**5.3 Summary of the study**

To further explore the robustness of the proposed model and its capacity to address small data samples, perturbations of 5%, 7% and 10% are applied to the data to validate the model's immunity to perturbations. Meanwhile, 20%, 30% and 50% of the data are used as training samples to verify the model's ability to handle small sample data.

In terms of robustness, the MSE is used as an accuracy metric because it better reflects the variation in model output. According to Table 15, the changes in the MSE for BRB1 under different perturbations are 0.0016, 0.0061, and 0.0083, and the changes in the MSE for BRB0 under different perturbations are 0.0027, 0.007, and 0.0096. The changes in BRB1 are all smaller than those in BRB0, which verifies that the BRBs with lower LCSs have stronger robustness. Additionally, it is verified that the model under the robustness constraint has stronger robustness. In addition, the MSE of BRB1 is smaller than that of LSTM, BPNN and RBF under different perturbations. As shown in Fig. 6, BRB1 has a significant advantage in the comparison of each model under different perturbations.

In terms of addressing small sample capacity data, MSE and accuracy serve as pivotal metrics for model assessment. Fig. 7 demonstrates that the BRB1 model maintains high accuracy consistently, even with limited training samples, highlighting its surperior in small sample contexts. In contrast, neural network models like LSTM, BPNN, and RBF perform poorly under similar conditions. This evidences the BRB1 model's superior capability to effectively handle small datasets, an essential trait for models operating in data-scarce environments.

Table 15: Comparison of MSE between BRB and other models with different disturbances

| | 0% | 5% | 7% | 10% |
|---|---|---|---|---|
| BRB0 | 0.0139 | 0.0166 | 0.0209 | 0.0235 |
| BRB1 | 0.0111 | 0.0127 | 0.0172 | 0.0194 |
| LSTM | 0.0151 | 0.0241 | 0.0292 | 0.0366 |
| BPNN | 0.0166 | 0.0242 | 0.0317 | 0.0393 |
| RBF | 0.0192 | 0.0242 | 0.0271 | 0.0288 |

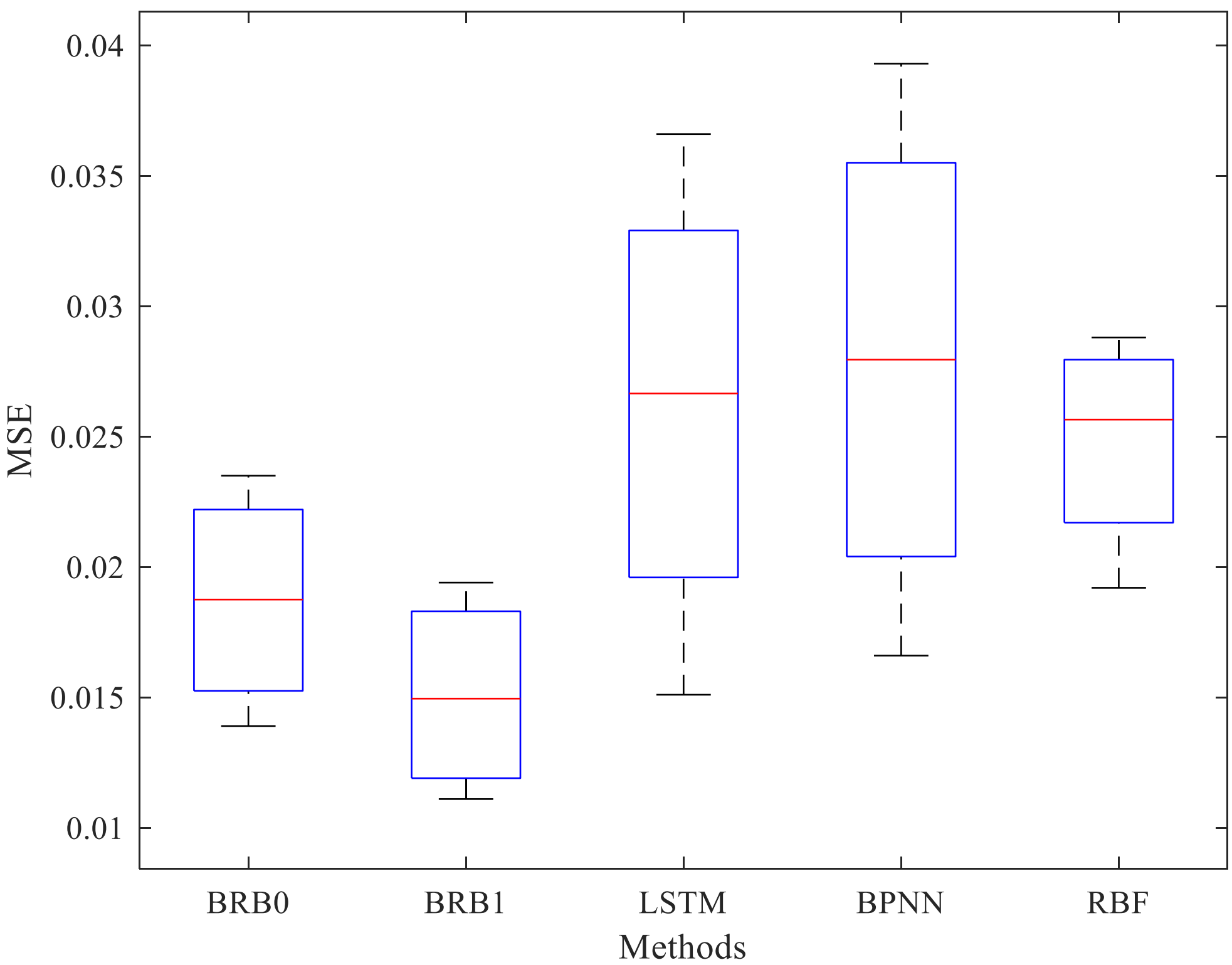

Fig. 6: Comparison of the stabilities of different methods

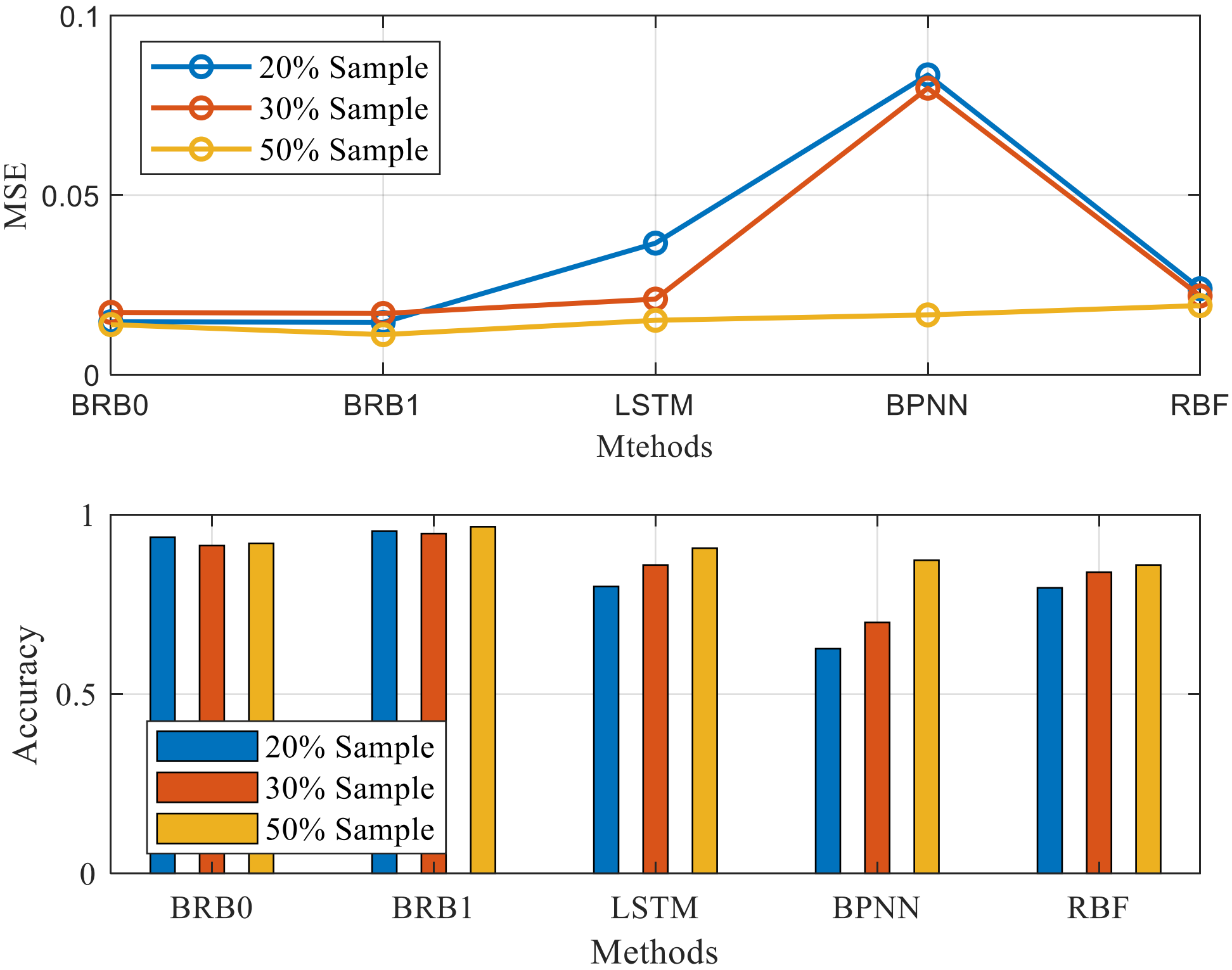

Fig. 7: Comparison of the accuracy of different methods with different samples

**6 Conclusion**

In this paper, we introduce a novel fault diagnosis method specifically designed for sensor failure detection in large equipment. This methodology fulfills essential demands for robustness assessment and enhancement. It features a detailed robustness analysis of the BRB and substantially improves model reliability through an optimized robustness constraint strategy.

Through detailed case studies involving the WD615 diesel engine and Case Western Reserve University bearing fault diagnosis, the proposed model demonstrates superior performance in terms of accuracy and robustness. Moreover, the model's generalization capability has been affirmed across diverse scenarios. However, while the method exhibits extensive applicability, it encounters limitations when dealing with more complex fault scenarios, such as those found in inertial navigation systems and servo mechanisms. These scenarios present unique challenges that our current model does not fully address, highlighting the need for tailored strategies in these specific contexts.

Future research will focus on overcoming these limitations by developing advanced diagnostic techniques that are adaptable to a broader range of systems. This will include extending our model to effectively handle the intricate fault diagnosis requirements of other critical applications, thereby broadening the scope and impact of our work.